\documentclass{article}
\usepackage{arxiv}
\usepackage{lineno}
\usepackage[utf8]{inputenc} 
\usepackage[T1]{fontenc}    
\usepackage{hyperref}       
\usepackage{url}            
\usepackage{booktabs}       
\usepackage{amsfonts}       
\usepackage{nicefrac}       
\usepackage{microtype}      
\usepackage{subcaption}
\usepackage{lipsum}
\usepackage{graphicx}
\graphicspath{ {./images/} }

\title{Explainable Flood Segmentation on Sentinel-1 SAR Imagery: A Comparative Study of CNN and Transformer Architectures}

\author{
 Arundhuti Banerjee \\
  United Nations University’s Institute for Environment and Human Security (UNU-EHS)\\
  Bonn, Germany 53113 \\
  \texttt{arundhuti14@gmail.com} \\
   \And
 David Daou\\
  United Nations University’s Institute for Environment and Human Security (UNU-EHS)\\
  Bonn, Germany 53113\\
  \texttt{daou@ehs.unu.edu} \\
}

\begin{document}
\maketitle
\begin{abstract}
Rapid and accurate flood prediction is essential for disaster response and mitigation planning. Synthetic Aperture Radar (SAR) sensors in satellites are well-suited for this purpose because they operate independently of weather and daylight conditions. Although SAR-based data enable all-weather flood monitoring, distinguishing flooded land from permanent water remains a significant challenge, particularly when flooding is defined strictly as inundated land. This study provides a comprehensive comparison of convolutional neural network (CNN) and vision transformer architectures for multi-class flood segmentation using Sentinel-1 SAR imagery, specifically trained to separate flooded land from permanent water bodies and land. Three state-of-the-art (SOTA)CNN-based models, U-Net, U-Net++, and DeepLabV3 with ResNet-34 backbone, and three SegFormer variants (b0, b1, and b2) were evaluated in two benchmark datasets, the ETCI NASA dataset and SenFloods11, using scene-based data splits to ensure a realistic assessment of spatial generalization. The results demonstrate that SegFormer-b2 significantly outperforms the U-Net baseline on the ETCI dataset (higher flood IoU across all 7 test scenes; Wilcoxon signed-rank), while after fine-tuning on Sen1Floods11, the advantage narrows to within the range of scene variability and is concentrated in spatially fragmented flood events. The study includes both qualitative and quantitative explainability techniques to visually comprehend model decisions and systematically assess prediction reliability. Qualitative analysis reveals that SegFormer-b2 produces more spatially coherent Grad-CAM activations focused on flood-relevant features, while U-Net generates more informative uncertainty estimates along flood boundaries. Quantitative analysis demonstrates that SegFormer-b2 achieves more faithful explanations and stronger entropy-error correlation, while U-Net provides more spatially precise explanations and marginally better calibration. These complementary strengths underscore the importance of integrating explainability analysis into operational SAR-based flood-monitoring systems. 
\end{abstract}
\section{Introduction}
Floods can cause widespread devastation, resulting in substantial economic and personal losses. According to the World Meteorological Organization (WMO), the frequency and intensity of flood events have increased significantly over the past few decades \cite{WMO}. This makes rapid and accurate flood mapping critical for disaster response. Satellite observations play a huge role in supporting the mitigation, response, and recovery phases of the disaster cycle. In particular, SAR (Synthetic Aperture Radar) sensors are important for flood detection because they are unaffected by clouds and operate independently of daylight conditions, making them reliable during extreme weather events \cite{Amitrano, Misra}. SAR images from the Sentinel-1 \cite{Bonafilia} mission have been widely used for flood prediction by utilizing the VV and VH polarization channels and thresholding the backscatter values of water surfaces \cite{Bonafilia, Twele, Liang2020a}. But this can be a problem in flood mapping, as flooded land from permanent water bodies exhibits similar backscatter characteristics \cite{Twele, Pulvirenti}. Distinguishing flooded land from permanent water bodies is essential for operational flood response because only newly inundated areas represent active hazards requiring emergency intervention.
Recent advancements in image processing techniques have significantly enhanced the analysis of remote sensing data, enabling more accurate and automated flood detection \cite{Bai, Bentivoglio, Ghosh, Portales, Wieland}. Deep learning approaches have shown promising results for flood segmentation, with convolutional neural networks (CNNs) widely adopted for their strong local feature extraction capabilities \cite{Wieland, Tavus}. Convolutional Neural Networks (CNNs), such as U-Net \cite{Ronneberger}, U-Net++ \cite{Zhou2018}, and DeepLabV3 \cite{Chen2017}, are widely used for flood segmentation using synthetic aperture radar (SAR) data  data due to their computational efficiency and effective extraction of local features. Although convolutional neural networks (CNNs) demonstrate strong performance, they frequently fail to produce satisfactory results when processing synthetic aperture radar (SAR) images with high variance in texture, shape, and feature representation \cite{Bentivoglio, Andrew, Saleh}. This limitation arises from the use of convolution operations, in which convolutional kernels focus only on neighboring pixels limiting its receptive field. As a result, CNNs cannot effectively capture long-range spatial dependencies, significantly reducing their performance on SAR-based flood segmentation tasks \cite{Andrew, Saleh}. 
Recently, transformer-based segmentation architectures have emerged as strong alternatives, providing improved representations of long-range spatial dependencies through self-attention mechanisms \cite{Yang, Luo, Hassija, Ugile}. However, systematic comparisons between CNNs and transformer-based architectures for SAR-based flood mapping, particularly under realistic and challenging labeling conditions, remain limited. Furthermore, most existing studies focus primarily on binary flood segmentation, in which models detect only water surfaces. Permanent water data is later substrated from the model predictions to highlight the temporary water surfaces.
Separating temporary water from permanent water is a more challenging task that requires the model to learn subtle differences in backscatter intensity, texture, and spatial context. In contrast, multi-class flood segmentation employs separate classes for flooded land and permanent water bodies, which remains comparatively underexplored despite its practical importance in real-world disaster scenarios \cite{Bharti, Kothari}. This study presents one of the first systematic explainability-focused comparisons of convolutional neural network (CNN) and transformer-based architectures for multiclass SAR flood segmentation, explicitly separating flooded land from permanent water bodies.  Specifically, the study evaluates CNN-based models and compares them with the hybrid vision-transformer-based SegFormer variants b0, b1, and b2. Furthermore, the analysis advances traditional performance evaluation by incorporating explainability and uncertainty estimation techniques. The main contributions of this paper are as follows: \par
(i)	Flood segmentation is formulated as a multiclass problem, in which flooded land is segregated from permanent water bodies,\par
(ii)	A comparative analysis of convolutional neural networks (CNNs) and vision transformer architectures to evaluate their ability to capture diverse flood features from SAR imagery,\par
(iii)	Flood performance is investigated using scene-based data splits across multiple datasets, enabling a more realistic assessment of spatial generalization and cross-dataset robustness,\par
(iv)	The qualitative and quantitative explainability aspects have been explored to provide insights into model predictions and their reliability.

\begin{figure}[htbp]
\centering
\includegraphics[width=0.5\columnwidth]{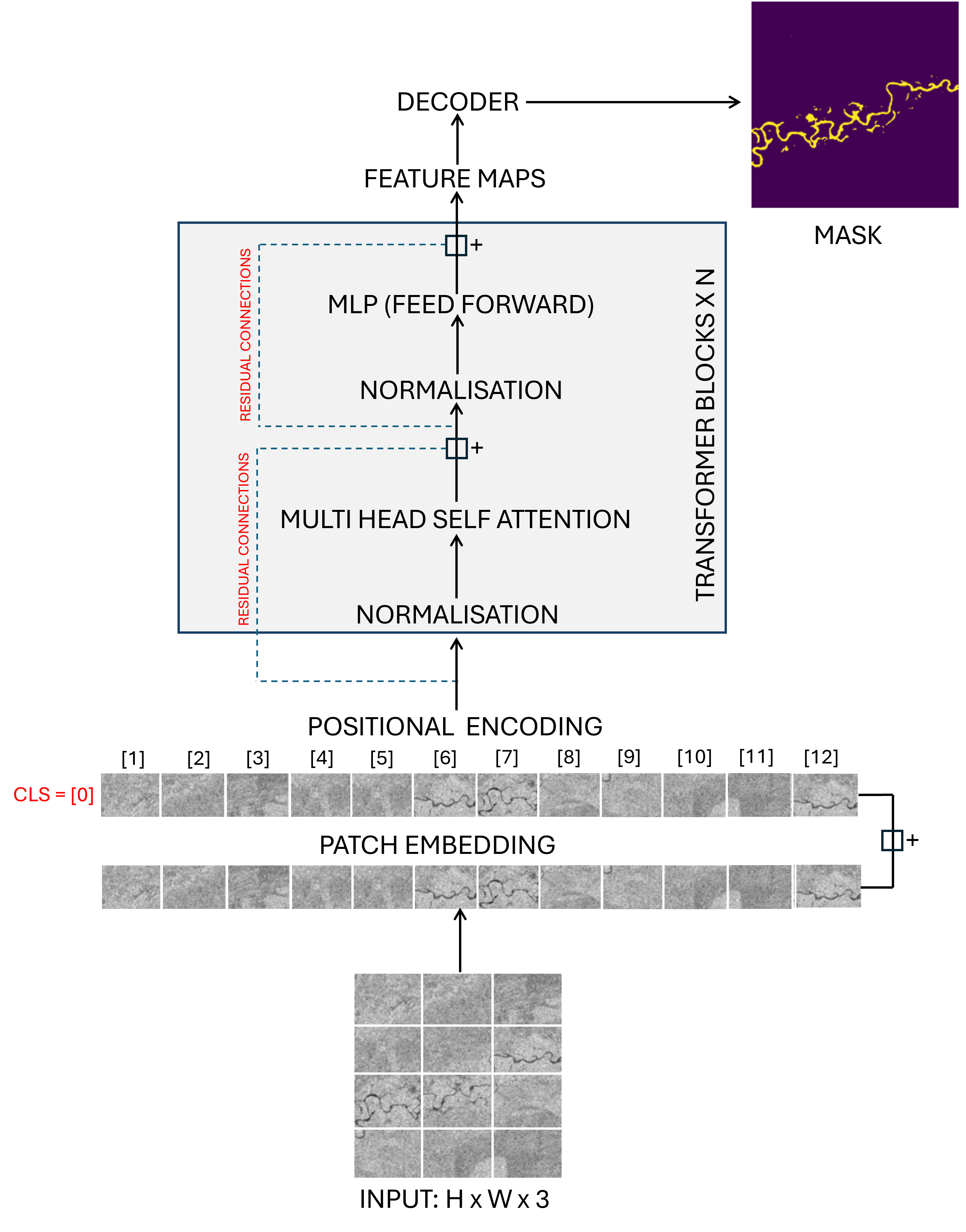}
\caption{Vanilla U-Net architecture with a symmetric encoder-decoder path, skip connections via concatenation, and a bottleneck in the middle.}
\label{fig:unet}
\end{figure}

\begin{figure}[htbp]
\centering
\includegraphics[width=0.5\columnwidth]{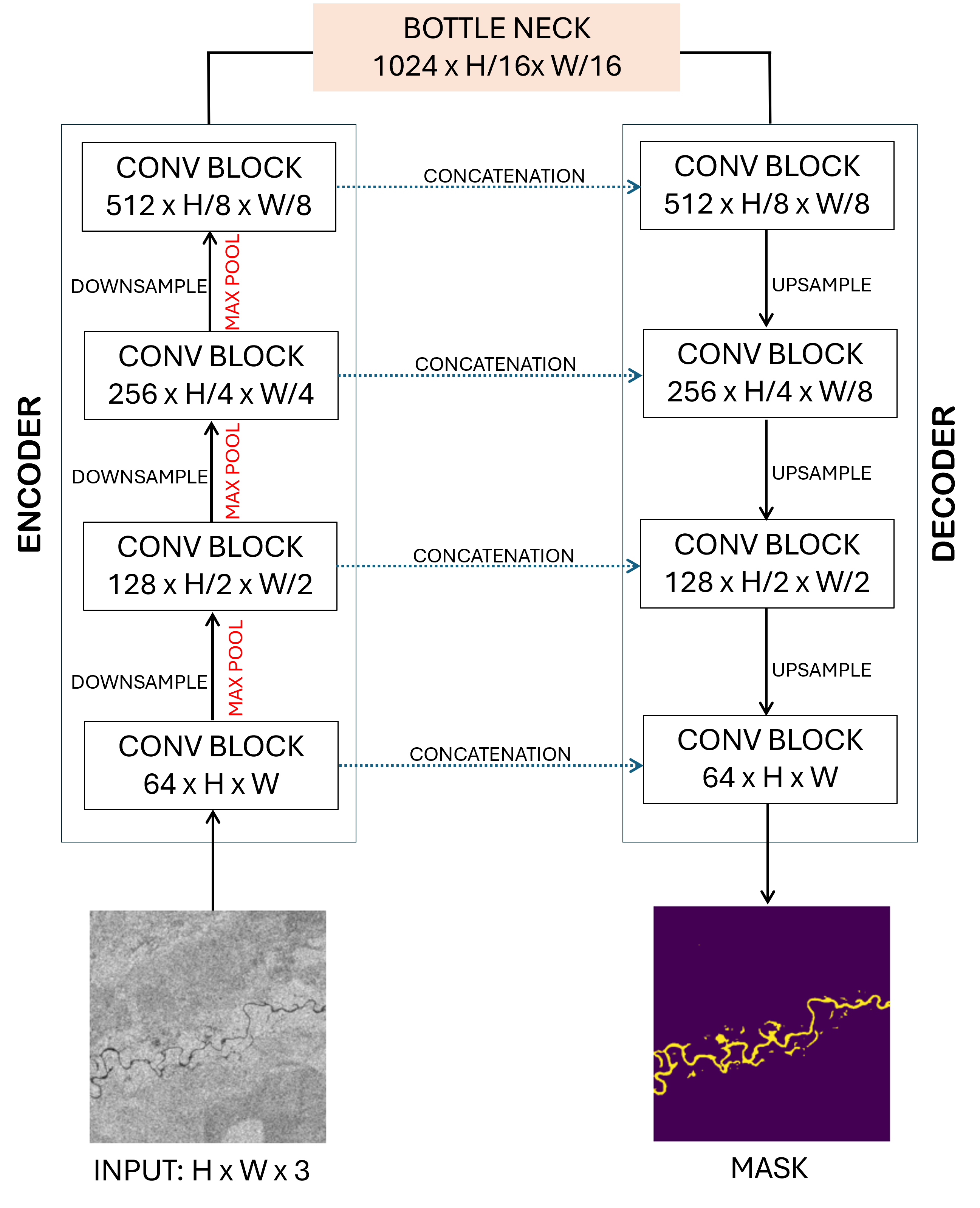}
\caption{Vanilla ViT encoder paired with a decoder for semantic segmentation.}
\label{fig:vit}
\end{figure}

\section{SAR-Based Flood Detection}
Synthetic Aperture Radar (SAR) has been widely used for flood detection due to its ability to operate under all-weather and all-light conditions and its capacity to penetrate clouds, which often pose challenges to optical sensors during flood events \cite{Bentivoglio, Tarpanelli}. SAR systems transmit microwave pulses and record the backscattered radiation that gets reflected from the Earth’s surface \cite{Gupta}. Permanent water bodies such as streams, rivers, and seas typically exhibit very low backscatter values in SAR imagery compared to surrounding land surfaces. These distinct radiometric characteristics lay the foundation for SAR-based flood mapping. Previous approaches primarily employed thresholding techniques that classify pixels as water based on a fixed backscatter intensity value \cite{Liang2020a, Guo2022}. Although computationally simple, these thresholding methods are highly sensitive to threshold value and often perform poorly in complex environments such as urban areas, vegetated floodplains, and regions with heterogeneous land cover \cite{Bentivoglio, Martinis}. Recent advancements in deep learning, particularly semantic segmentation algorithms, have significantly improved the accuracy and robustness of flood mapping from SAR imagery \cite{Bentivoglio, Ghosh, Portales, Wieland, Tavus}. Among these, the U-Net architecture, originally developed for biomedical image segmentation \cite{Ronneberger}, has been widely used for SAR flood mapping \cite{Bai, Ghosh, Paul}. Its symmetric encoder-decoder structure with skip connections effectively preserves fine spatial details, which is important for capturing irregular flood boundaries \cite{Wang2025}. Several studies have used the U-Net and its variants to demonstrate their effectiveness in flood detection. Paul and Ganju 2021 \cite{Paul} demonstrated the effectiveness of semi-supervised techniques using U-Net and U-Net++ ensembles for SAR flood segmentation. Wu et al. (2022) \cite{Wu2022} used a Deeplab v3++ model with a dual-channel MobileNetV2 backbone, achieving strong boundary delineation in dual-polarization SAR imagery. Li et al. (2023) \cite{Li} proposed a novel approach that combines the U-Net architecture with Google Earth Engine (GEE) to enable rapid and accurate flood mapping using Sentinel-1 SAR data. Ghosh et al 2024 \cite{Ghosh} used a nested U-Net (U-Net++) architecture with EfficientNet-B7 as the encoder backbone for automatic flood segmentation. Wu et al. (2023) \cite{Wu2023} compared various U-Net models and focused on developing an operational, real-time flood detection pipeline. Similarly, Karci et al. \cite{Karci} systematically evaluated the performance of three segmentation architectures, U-Net, SegNet, and DeepLabV3+, using different backbones for flood segmentation. Although Convolutional Neural Networks (CNNs) have been widely employed for flood mapping, they exhibit inherent limitations in SAR-based flood segmentation. Due to their dependence on small, fixed-size convolutional kernels, CNNs have a limited receptive field, which limits their ability to effectively capture long-range spatial dependencies and large-scale contextual information across extensive flood regions \cite{Saleh, Noori}.
Vision Transformers (ViTs) overcome these limitations by leveraging self-attention mechanisms that effectively capture global and local contextual information across the entire image \cite{Yang}.  Saleh et al. (2024) \cite{Saleh} proposed a vision transformer-based architecture for bi-temporal SAR flood detection and demonstrated improved handling of speckle noiJamali et al. (2024) \cite{Jamali} combined the UNet architecture with a Vision Multi-Layer Perceptron (MLP) and reported a noticeable improvement in capturing global and local features in SAR images.ges. Similarly, Zhou et al. 2025 \cite{Zhou2025} incorporated transformer attention modules into U-Net architectures to better capture SAR-specific features. Sharma et al. (2025) \cite{Sharma} demonstrated that ViT-based deep ensembles significantly outperform traditional CNN-only models for SAR-based flood inundation mapping. Among hybrid transformer architectures, SegFormer is widely recognized for its lightweight design, computational efficiency, and ability to learn complex features through its hierarchical transformer encoder and simple decoder structure \cite{Xie}. Zou et al. \cite{Zou} combined a SegFormer model with a convolution-based U-Net architecture to address SAR-specific challenges such as speckle. Chen et al. \cite{Chen2026} evaluated several binary flood segmentation architectures, including U-Net, U-Net++, DeepLabV3+, and SegFormer, for flood detection on both optical and SAR data. Although SegFormer demonstrated competitive performance, U-Net++ outperformed it, achieving higher overall segmentation accuracy and robustness across complex scenarios.
Despite the success of modern deep learning models in SAR flood segmentation, their black-box nature poses significant challenges to their practical application. In real disaster management situations, emergency responders and decision-makers seek transparent explanations for reliable predictions. Consequently, recent research has highlighted the importance of integrating Explainable AI (XAI) techniques to improve model transparency, interpretability, and operational reliability \cite{Yu, Rezvani, Shirmohammadi, Tosan}. Sanderson et al. \cite{Sanderson} used Class Activation Mapping (CAM)-based techniques, particularly Grad-CAM and HiRes-CAM, to visualize the regions that influenced the model’s flood predictions. Ghosh et al. \cite{Ghosh} and Bathe and Patil (2025) \cite{Bathe} used SHAP-based visualizations to understand which regions in the images most strongly influence flood prediction outcomes. Yu et al. 2026 \cite{Yu} used an explainability technique to visualize important regions in the model’s decision-making for flood inundation mapping with Sentinel-1, Sentinel-2, and DEM inputs. 

\section{Flood Dataset}
Two benchmark datasets have been utilized in this study to train and evaluate the flood segmentation models. These datasets provide diverse geographic coverage, thereby enabling a detailed assessment of model performance. 
\subsection{ETCI Dataset}
The primary dataset used to train the model from scratch is from the NASA Disaster Team's flood detection challenge \cite{Paul}. This ETCI dataset consists of 66,000 tiled 256 x 256 SAR images from varied geographic regions, including North Alabama (US), Bangladesh, and Nebraska (US). Each tile is derived from Sentinel-1 C-band synthetic aperture radar (SAR) imagery acquired in interferometric wide-swath mode at 5-meter-by-20-meter resolution, with dual polarization (VV and VH). Additionally, a third channel, derived as $1 - (VH/VV)$, is stacked with the VV and VH channels, thereby enhancing the contrast between water and non-water surfaces \cite{Paul, Bonafilia}. The ETCI dataset is heavily imbalanced, with significantly fewer flood pixels than the background and permanent water classes. To address this issue, a hybrid loss function combining Dice loss and Focal loss function is used to improve sensitivity to minority classes. To mitigate class imbalance at the tile level, a balanced sampling strategy with upsampling was employed, in which tiles containing at least 600 flood pixels were treated as positive samples. The ETCI dataset comprises 33 scenes, where each scene corresponds to one
Sentinel-1 acquisition (one region on one date). To avoid spatial leakage, the split was performed at the scene level: entire scenes were randomly assigned to training (19 scenes, 20{,}665 tiles), validation (7 scenes, 6{,}872 tiles), and test subsets (7 scenes, 7{,}948 tiles) so that no scene has overlapping tiles. The resulting test set comprises seven scenes covering three geographic regions: Bangladesh, Nebraska, and North Alabama. Quantitative evaluation on the ETCI  dataset is performed on the 940 test tiles containing at least 600 flood pixels which is consistent with the flood-presence requirement of the explainability analyses.
\subsection{11 Dataset}
The trained segmentation model was then further fine-tuned on the Sen1Floods11 flood dataset \cite{Bonafilia}. The Sen1Floods11 dataset comprises 4831 image tiles, each 512 × 512 pixels, covering a total area of 120406 km² across various geographic regions worldwide. Of the 4831 chips, 4370 were automatically annotated using simple remote sensing classification algorithms and used as weakly supervised training data. The remaining 446 chips were hand-labeled and used as high-quality training, testing, and validation data. More details are provided in \cite{Bonafilia}. In the present study, only the hand-labeled data were used for the analysis. The hand labelled chips were split as: 251 chips for fine-tuning/training, 89 for validation, and a fixed test set of 28 chips drawn from the official test split, covering six flood events across six regions: India (10 chips), Paraguay (7), Spain (5), Ghana (3), the Mekong basin (2), and Nigeria (1).Since the hand labels provide a binary water mask, three-class targets were constructed by combining them with the dataset's JRC permanent-water layer: permanent-water pixels were assigned to class~1, hand-labeled water outside the permanent-water mask to class~2 (flooded land), and invalid pixels were excluded from training and evaluation. Quantitative evaluation is performed on the 27 test chips containing flood pixels; one chip without flood pixels was excluded.
\section{Experimental Setup and Results}
\subsection{Training Configuration}
The main aim of the study is to train a multi-class segmentation model with 3 classes: class 0 (background), class 1 (permanent water), and class 2 (flooded land). Initial training on the ETCI dataset was performed for up to 100 epochs with a batch size of 16, followed by fine-tuning on the Sen1Floods11 dataset for up to 40 epochs. Mixed precision training was implemented. The model was optimized using the validation flood Intersection over Union (IoU) metric as the early-stopping criterion to improve the classification of flooded land. The AdamW optimizer was used with a suitable weight decay of $10^{-4}$. A combined loss function comprising Dice loss and Focal loss ($\gamma$=2.0) was implemented to address class imbalance and improve segmentation quality. Early stopping with a patience of 8 epochs was applied to reduce overfitting. For fine-tuning on the Sen1Floods11 dataset, all parameters remained unchanged except for the batch size, which was set to 8. For this study, U-Net, U-Net++, and DeepLabV3 were employed as representative convolutional neural network (CNN)-based segmentation architectures. For vision transformer-based models, SegFormer variants b0, b1, and b2 have been used. For brevity, the SegFormer variants are referred to as SegFormer-b0, SegFormer-b1, and SegFormer-b2 throughout this paper, although SegFormer is a hybrid architecture that combines a hierarchical transformer encoder with a lightweight MLP decoder and is distinct from the vanilla Vision Transformer. Except for SegFormer, all models used ResNet34 as the backbone. The learning rate for UNet models was set to $10^{-4}$, while for SegFormer and DeepLabV3 it was set to $5\times 10^{-5}$. 

\subsection{Evaluation Metrics}
Segmentation performance was evaluated using standard metrics, including accuracy, precision, F1 score, and intersection over union (IoU). These metrics were computed separately for permanent water (class 1) and flood on land (class 2) segmentation.\par
\textbf{Intersection over Union (IoU)} is defined as the ratio of the overlapping area between the predicted pixels and the ground truth pixels to the total area covered by both. This metric quantifies the degree of alignment between the predicted segmentation and the true labels for each class.
\begin{equation}
\label{deqn_ex1}
 \frac{TP}{TP+FP+FN}
\end{equation}
\textbf{Precision} is defined as the ratio of correctly predicted positive instances to the total number of positive instances predicted. This metric evaluates the model’s capacity to reduce false triggers by minimizing false positives.
\begin{equation}
\label{deqn_ex2}
 \frac{TP}{TP+FP}
\end{equation}
\textbf{Recall} is defined as the ratio of correctly predicted positive instances to the total number of actual positive instances. This metric assesses the model’s ability to identify all relevant instances.
\begin{equation}
\label{deqn_ex3}
 \frac{TP}{TP+FN}
\end{equation}
\textbf{F1 score} is defined as the harmonic mean of precision and recall.
\begin{equation}
\label{deqn_ex4}
 \frac{2\times Precison\times Recall}{Precison+Recall}
\end{equation}
\textbf{Statistical Significance Analysis} determines whether performance differences between different architectures exceed scene-to-scene variability. Paired statistical analyses were performed on the per-scene (ETCI dataset) and per-chip (Sen1Floods11 dataset) flood IoU of the architectures. The unit of statistical independence was chosen according to the spatial structure of corresponding datasets: for ETCI, it was the regional scene, since tiles within a scene are croppings of a single acquisition and are therefore spatially correlated. For Sen1Floods11, it was the individual chip, since chips correspond to geographically distinct locations. For each unit, a confusion matrix was calculated separately and the flood IoU was derived from it. Differences were tested with the two-sided paired Wilcoxon signed-rank test~\cite{wilcoxon1945, demsar2006}, and the uncertainty of the mean difference was quantified with a bootstrap 95\% confidence interval (10{,}000 resamples over scenes/chips).
\subsection{Explainability}
In this study, the reliability of flood segmentation predictions is assessed using explainability techniques. For qualitative analysis, entropy maps \cite{Shannon} and Gradient-weighted Class Activation Mapping (Grad-CAM) maps \cite{Selvaraju} are employed. Entropy maps generated from the predicted class probabilities quantify pixel-level uncertainty by calculating the entropy of the SoftMax output distribution. Grad-CAM maps are calculated using gradients from the model's final convolutional layers that contribute most to its predictions. In flood segmentation, these activation maps indicate whether the model pays attention to physically meaningful features, such as water boundaries and inundated regions, or is influenced by irrelevant patterns. This approach is essential for evaluating model behavior in challenging scenarios, including overlapping flood and permanent-water regions. For quantitative analysis, the following statistics were used: \par
\textbf{1. Expected Calibration Error (ECE)} quantifies the difference between a model's predicted confidence and its actual accuracy \cite{guo2017calibration, naeini2015obtaining}.  A model is considered confidence-calibrated if, for all confidences, the model is correct a proportion of the time:
\begin{equation}
\label{ECE}
ECE = \sum_{m=1}^{M} \frac{|B_m|}{n} |acc(B_m) - conf(B_m)|,
\end{equation}
where $M$ is the number of confidence bins, $B_m$ is the set of pixels in bin $m$, $n$ is the total number of pixels, $acc(B_m)$ is the fraction of correctly classified pixels in bin $m$, and $conf(B_m)$ is the mean predicted confidence in bin $m$.\par
\textbf{2. Entropy–Error Correlation} quantifies the relationship between pixel-wise predictive entropy and pixel-wise prediction error using the Pearson correlation coefficient. Predictive entropy is computed from the softmax output distribution as:
\begin{equation}
\label{entropy}
H=-\sum{p(c)} \log{p(c)}
\end{equation}
\textbf{3. Deletion and Insertion Faithfulness} quantifies whether the Grad-CAM explanation correctly detects the image features important to the model's prediction \cite{petsiuk2018rise}. In the deletion test, pixels are removed from the input image in order of decreasing Grad-CAM saliency, and the area under the resulting confidence degradation curve (AUDC) is computed. A lower AUDC indicates a more faithful explanation. In the insertion test, pixels are added to a blank baseline input in decreasing order of saliency, and the area under the confidence recovery curve (AUIC) is computed; a higher AUIC indicates that salient features are sufficient to restore model performance.\par
\textbf{4. Spatial Entropy Analysis} quantifies the meaningfulness of the model's uncertainty by computing mean predictive entropy \cite{Shannon} separately across three semantic zones: boundary pixels, interior flood pixels, and interior non-flood pixels. This decomposition reveals whether the model is appropriately uncertain at class boundaries and in ambiguous flood regions versus confident in unambiguous areas.\par
\textbf{5. The pointing game} quantifies the spatial precision of Grad-CAM explanations by checking whether the single pixel with the highest activation lies within the ground-truth flood region \cite{zhang2018topdown}. The score is the ratio of the most salient pixel within the flood mask to the total number of test samples. 
\subsection{Test Time Augmentation (TTA)}
Test-Time Augmentation (TTA) is a widely used technique that uses data augmentation at inference to improve prediction robustness \cite{Wang2019}.In TTA, multiple predictions from transformed versions of the same input are used and aggregated. This mitigates the model's sensitivity to specific orientations and intensity variations, producing more stable and reliable segmentation outputs. This is particularly useful for flood mapping from SAR imagery, where backscatter values can exhibit high variance due to environmental differences \cite{Twele, Chini}. In this study, geometric transformations, specifically horizontal and vertical flips, are combined with multiplicative intensity scaling factors (0.95, 1.0, and 1.05) to simulate minor variations in synthetic aperture radar (SAR) backscatter. These transformations yield 12 augmented versions of each input image. For each augmented input, the model produces class-wise logits.The final prediction is obtained by averaging the logits over all augmented inputs before class selection. 
\section{Results}
\subsection{ETCI Dataset Results}
The model was trained and validated on the ETCI training for 100 epochs. Table I presents the flood-on-land (class 2) segmentation performance of the model on the ETCI dataset. The results show that, among all architectures evaluated, the transformer-based model SegFormer-b2 performs best, with a flood IoU of 0.313 and an F1-score of 0.477. Compared to CNN-based models such as UNet, UNet++, and DeepLabV3, SegFormer demonstrates improved precision and recall, indicating a better balance between detecting flooded regions and avoiding false positives. Most previous studies treat flood detection as a binary segmentation problem, where permanent water and flooded areas are merged into a single class. In contrast, this study addresses a more challenging multi-class task by explicitly distinguishing flooded land from permanent water bodies, which explains the low segmentation score. This formulation is more relevant for operational flood response, where identifying newly inundated areas is critical. Table II presents a similar table for permanent water class segmentation; in this model, the class is 1.  In contrast to flood detection, DeepLabV3 achieves the highest water IoU, demonstrating that CNN-based architectures are highly effective at segmenting homogeneous, well-defined water bodies. It needs to be highlighted again that flood pixels overlapping permanent water are reassigned to the water class, making flood detection more challenging and further emphasizing the importance of the model's capability to distinguish subtle differences. Also, the model training criteria are to perform better on the flood-on-land class.  The goal of the study is to improve the segmentation model's performance for the flood class.

\begin{table}[htbp]
\caption{Flood class segmentation performance across different model architectures on the ETCI dataset}
\label{tab:flood_results}
\centering
\begin{tabular}{l c c c c}
\toprule
\textbf{Model} & \textbf{IoU} & \textbf{Recall} & \textbf{Precision} & \textbf{F1} \\
\midrule
UNet (resnet34)      & 0.248 & 0.596 & 0.299 & 0.398 \\
UNet++ (resnet34)    & 0.229 & 0.590 & 0.272 & 0.373 \\
DeepLabV3 (resnet34) & 0.273 & 0.674 & 0.315 & 0.429 \\
\midrule
SegFormer-b0         & 0.241 & 0.588 & 0.291 & 0.389 \\
SegFormer-b1         & 0.295 & 0.656 & 0.348 & 0.455 \\
SegFormer-b2         & \textbf{0.313} & \textbf{0.726} & \textbf{0.355} & \textbf{0.477} \\
\bottomrule
\end{tabular}
\end{table}

\begin{table}[htbp]
\caption{Performance of the multi-class segmentation model on the permanent water class on the ETCI test dataset}
\label{tab:water_results}
\centering
\begin{tabular}{l c c c c}
\toprule
\textbf{Model} & \textbf{IoU} & \textbf{Recall} & \textbf{Precision} & \textbf{F1} \\
\midrule
UNet (resnet34)      & 0.663 & 0.721 & 0.892 & 0.797 \\
UNet++ (resnet34)    & 0.657 & 0.725 & 0.874 & 0.792 \\
DeepLabV3 (resnet34) & 0.749 & 0.810 & 0.909 & 0.856 \\
\midrule
SegFormer-b0         & 0.665 & 0.723 & 0.892 & 0.799 \\
SegFormer-b1         & 0.697 & 0.750 & 0.908 & 0.821 \\
SegFormer-b2         & \textbf{0.737} & \textbf{0.779} & \textbf{0.930} & \textbf{0.848} \\
\bottomrule
\end{tabular}
\end{table}

\begin{table}[htbp]
\caption{Performance of the multi-class segmentation model on the flood class (class 2) on the Sen1Floods11 test dataset}
\label{tab:sen1flood_flood}
\centering
\begin{tabular}{l c c c c}
\toprule
\textbf{Model} & \textbf{IoU} & \textbf{Recall} & \textbf{Precision} & \textbf{F1} \\
\midrule
\multicolumn{5}{c}{\textit{Without Test Time Augmentation}} \\
\midrule
UNet (resnet34)      & 0.408 & 0.561 & 0.598 & 0.579 \\
UNet++ (resnet34)    & 0.402 & 0.552 & 0.596 & 0.573 \\
DeepLabV3 (resnet34) & 0.382 & 0.545 & 0.564 & 0.553 \\
\midrule
SegFormer-b0         & 0.388 & 0.555 & 0.563 & 0.559 \\
SegFormer-b1         & 0.409 & 0.563 & 0.600 & 0.581 \\
SegFormer-b2         & \textbf{0.418} & \textbf{0.565} & \textbf{0.615} & \textbf{0.589} \\
\midrule
\multicolumn{5}{c}{\textit{With Test Time Augmentation}} \\
\midrule
UNet (resnet34)      & 0.419 & 0.564 & 0.620 & 0.591 \\
UNet++ (resnet34)    & 0.412 & 0.566 & 0.602 & 0.584 \\
DeepLabV3 (resnet34) & 0.391 & 0.541 & 0.585 & 0.562 \\
\midrule
SegFormer-b0         & 0.396 & 0.553 & 0.582 & 0.567 \\
SegFormer-b1         & 0.417 & 0.573 & 0.605 & 0.589 \\
SegFormer-b2         & \textbf{0.424} & \textbf{0.564} & \textbf{0.629} & \textbf{0.595} \\
\bottomrule
\end{tabular}
\end{table}

\begin{table}[htbp]
\caption{Performance of the multi-class segmentation model on the water class (class 1) on the Sen1Floods11 dataset}
\label{tab:sen1flood_water}
\centering
\begin{tabular}{l c c c c}
\toprule
\textbf{Model} & \textbf{IoU} & \textbf{Recall} & \textbf{Precision} & \textbf{F1} \\
\midrule
\multicolumn{5}{c}{\textit{Without Test Time Augmentation}} \\
\midrule
UNet (resnet34)      & 0.541 & 0.633 & 0.789 & 0.702 \\
UNet++ (resnet34)    & 0.570 & 0.652 & 0.820 & 0.726 \\
DeepLabV3 (resnet34) & 0.516 & 0.638 & 0.730 & 0.681 \\
\midrule
SegFormer-b0         & \textbf{0.595} & \textbf{0.668} & 0.845 & \textbf{0.746} \\
SegFormer-b1         & 0.583 & 0.661 & 0.830 & 0.736 \\
SegFormer-b2         & 0.583 & 0.642 & \textbf{0.863} & 0.736 \\
\midrule
\multicolumn{5}{c}{\textit{With Test Time Augmentation}} \\
\midrule
UNet (resnet34)      & 0.579 & 0.667 & 0.814 & 0.733 \\
UNet++ (resnet34)    & 0.570 & 0.637 & 0.845 & 0.726 \\
DeepLabV3 (resnet34) & 0.530 & 0.645 & 0.748 & 0.693 \\
\midrule
SegFormer-b0         & \textbf{0.628} & \textbf{0.698} & 0.863 & \textbf{0.772} \\
SegFormer-b1         & 0.582 & 0.647 & 0.853 & 0.736 \\
SegFormer-b2         & 0.614 & 0.679 & \textbf{0.866} & 0.761 \\
\bottomrule
\end{tabular}
\end{table}

\subsection{Sen1Floods11 Results}
The model initially trained on the ETCI dataset was subsequently fine-tuned on the Sen1Floods11 dataset for 40 epochs to facilitate cross-domain generalization. Test time augmentation was also applied to evaluate the model's performance. Table III presents the performance of all models evaluated for the flood-on-land (class 2) detection task on the Sen1Floods11 test dataset. In comparison to ETCI, flood class segmentation performance on the Sen1Floods11 dataset is higher, with IoU values ranging from 0.38 to 0.42 and F1 scores from 0.55 to 0.59. SegFormer-b2 achieves the highest performance, with a flood IoU of 0.418 without test-time augmentation, although its advantage over CNN models is only marginal. CNN-based models such as U-Net and DeepLabV3 achieve competitive performance but have slightly lower scores. This difference may stem from their reliance on local receptive fields, which can limit their ability to capture long-range spatial dependencies in flood regions, whereas transformer-based architectures are better suited to capturing complex, spatially fragmented flood patterns. The precision-recall balance in the Sen1Floods11 dataset is also more favorable than in the ETCI dataset. Test-time augmentation also yields only limited improvement for most models, suggesting that they are already robust to geometric transformations.
\begin{figure*}[t]
\centering
\includegraphics[width=\textwidth]{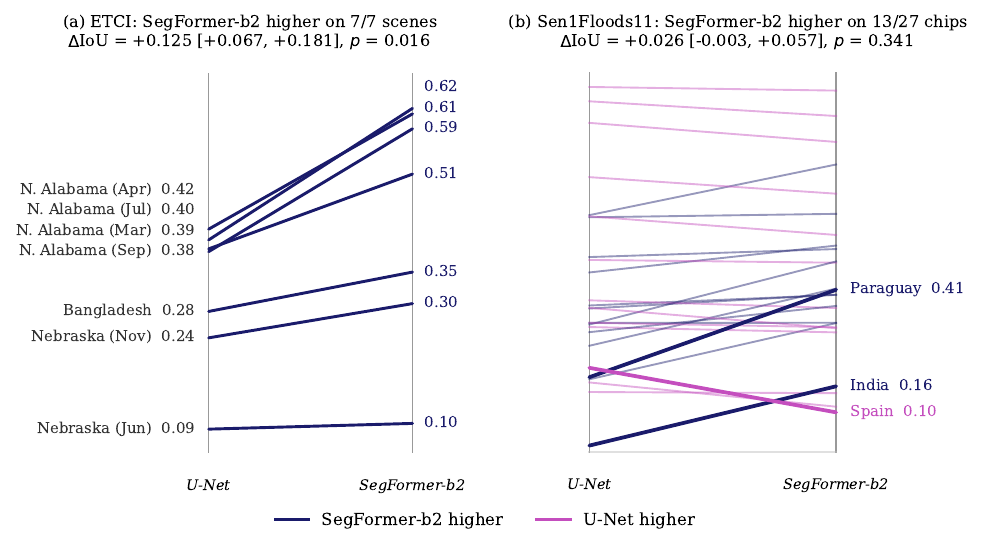}
\caption{Paired per-scene (a) and per-chip (b) flood IoU comparison
between U-Net and SegFormer-b2; panel (b) shows the result without test-time augmentation (TTA).}
\label{fig:slope}
\end{figure*}
To assess whether the observed differences exceed scene-to-scene variability, paired analyses were performed as described in Section~IV-B. On the ETCI test split, SegFormer-b2 achieved a higher flood IoU than U-Net on all 7 of 7 test scenes (Figure~\ref{fig:slope}a), with a mean per-scene advantage of $+0.125$
(95\% bootstrap CI $[+0.067, +0.181]$; Wilcoxon signed-rank
$p = 0.016$) -- a statistically significant difference that holds in
all three geographic regions of the test split. On the Sen1Floods11
test set, by contrast, the per-chip advantage was $+0.026$ without test time augmentation (TTA) and $+0.024$ with TTA, not statistically significant in either condition ($p = 0.34$ and $p = 0.41$; wins split 13--13 and 14--12,
median difference $\approx 0$; Figure~\ref{fig:slope}b), and concentrated
in the two events with fragmented, spatially complex inundation (India,
Paraguay), driven by chips on which U-Net fails substantially (e.g.,
0.016 vs.\ 0.165 IoU). 
Note that per-scene averaging weights every flood event equally, whereas Tables~I and~III pool pixels globally and
are therefore dominated by the largest events. Together, these statistics
indicate that SegFormer's advantage is robust when both architectures are trained from a common starting point on ETCI. However, it narrows to within-scene-to-scene variability after fine-tuning on Sen1Floods11, with the residual advantage appearing primarily in spatially fragmented flood patterns.
Table IV presents the segmentation performance of all models for the water class (class 1) in the Sen1Floods11 dataset. The results demonstrate that the model's ability to segment permanent water has decreased on the Sen1Floods11 dataset compared to its performance on the ETCI dataset. One possible explanation is that Sen1Floods11 contains more varied water labels, or permanent water bodies in this dataset are more fragmented, making them more challenging to segment. Without test-time augmentation (TTA), the SegFormer variants achieve intersection over union (IoU) scores ranging from 0.583 to 0.595, compared to 0.516 to 0.570 for the convolutional neural network (CNN) baselines. This trend is also observed in F1 scores: SegFormer models achieve 0.736-0.746, while CNN models achieve 0.681-0.726.  Across all models, precision scores (0.730-0.866) are consistently higher than recall scores (0.633-0.698), indicating that the models reliably avoid false positives. TTA yields meaningful performance improvements across most models, although the magnitude of improvement varies by architecture. The greatest improvements are observed for UNet, SegFormer-b0, and SegFormer-b2. In particular, the smallest SegFormer variant (b0) outperforms its larger counterparts (b1 and b2), a result that holds with and without TTA. These findings indicate that transformer-based models provide superior performance for flood detection, while CNNs remain competitive for structured regions, such as permanent water. The improvements achieved through TTA further highlight the importance of robustness in SAR-based flood mapping. \par

\begin{figure*}[t]
\centering
\begin{subfigure}[t]{0.28\textwidth}
    \centering
    \includegraphics[width=\linewidth]{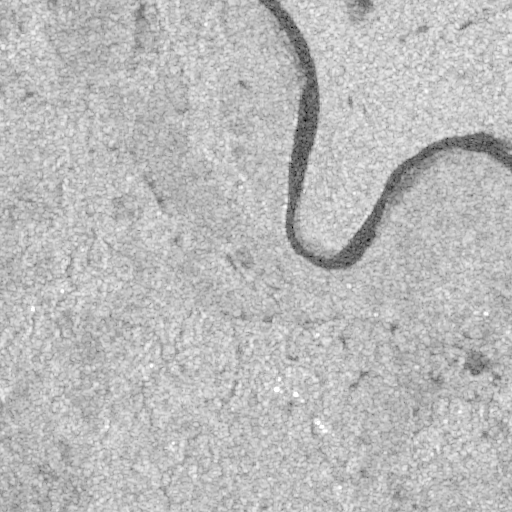}
    \caption{}
    \label{fig:scene1_vv}
\end{subfigure}
\hfill
\begin{subfigure}[t]{0.28\textwidth}
    \centering
    \hspace{-1.1cm}\includegraphics[width=\linewidth]{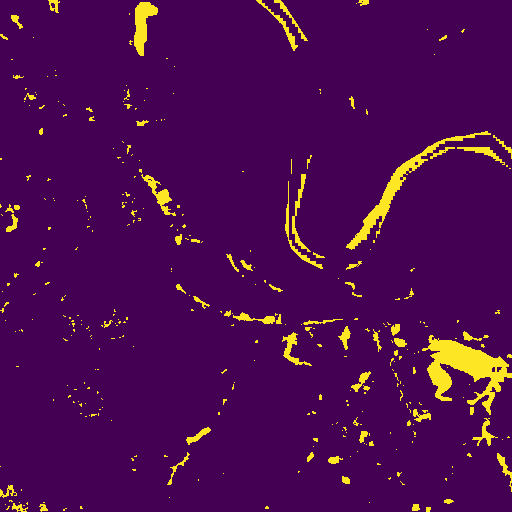}
    \caption{}
    \label{fig:scene1_flood_gt}
\end{subfigure}
\hfill
\begin{subfigure}[t]{0.28\textwidth}
    \centering
    \hspace{-1.1cm}\includegraphics[width=\linewidth]{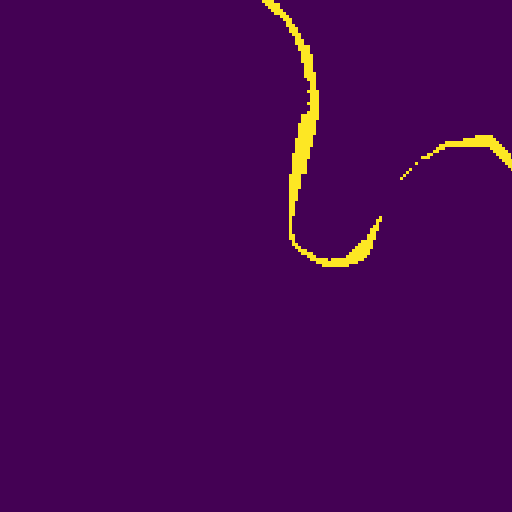}
    \caption{}
    \label{fig:scene1_water_gt}
\end{subfigure}

\vspace{0.3cm}

\begin{subfigure}[t]{0.28\textwidth}
    \centering
    \includegraphics[width=\linewidth]{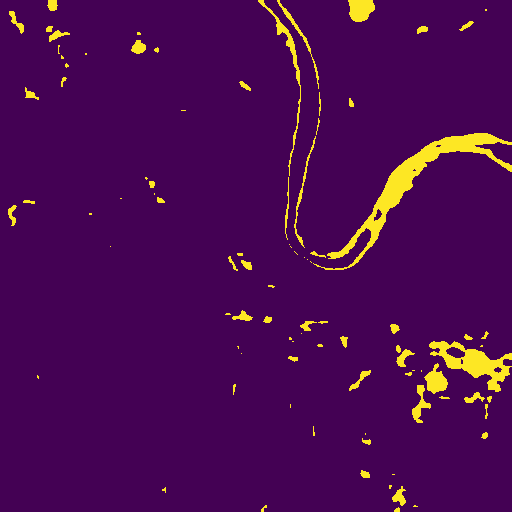}
    \caption{}
    \label{fig:scene1_segformer_prediction}
\end{subfigure}
\hfill
\begin{subfigure}[t]{0.33\textwidth}
    \centering
    \includegraphics[width=\linewidth]{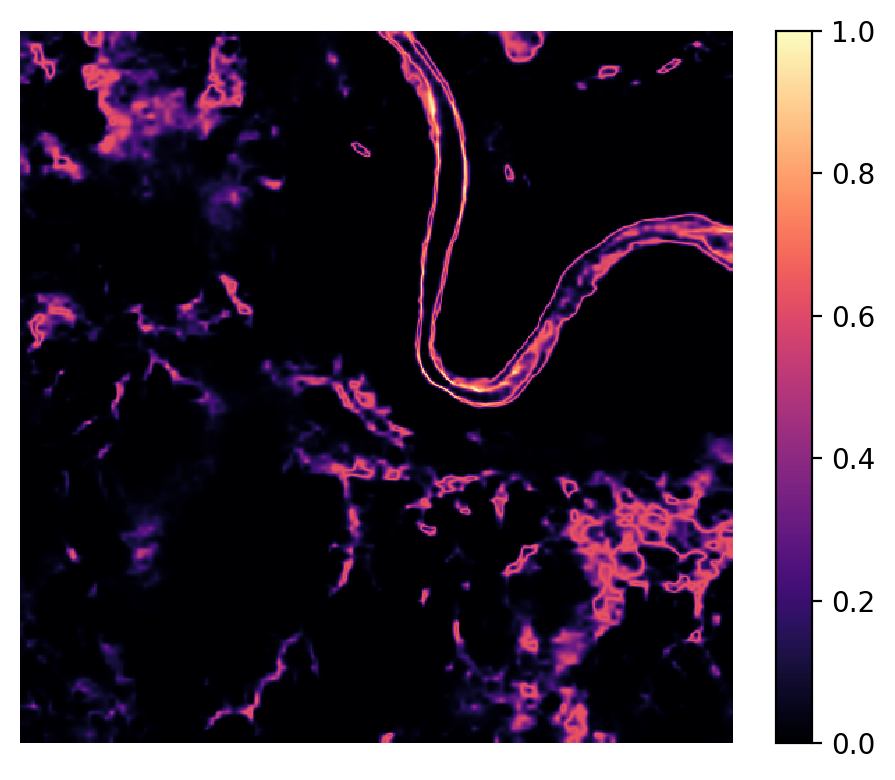}
    \caption{}
    \label{fig:scene1_segformer_entropy}
\end{subfigure}
\hfill
\begin{subfigure}[t]{0.33\textwidth}
    \centering
    \includegraphics[width=\linewidth]{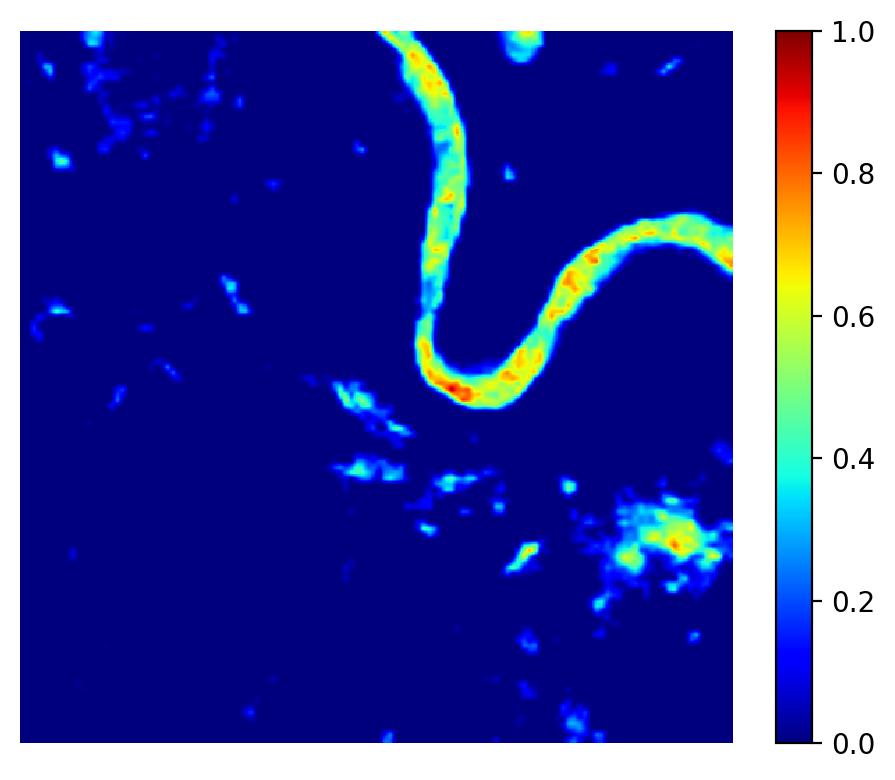}
    \caption{}
    \label{fig:scene1_segformer_gradcam}
\end{subfigure}

\vspace{0.3cm}

\begin{subfigure}[t]{0.28\textwidth}
    \centering
    \includegraphics[width=\linewidth]{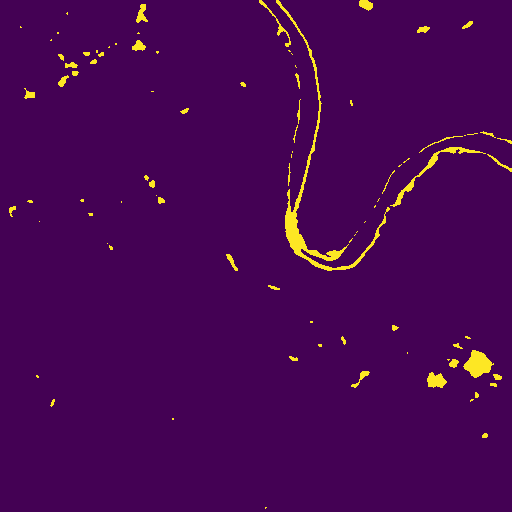}
    \caption{}
    \label{fig:scene1_unet_prediction}
\end{subfigure}
\hfill
\begin{subfigure}[t]{0.33\textwidth}
    \centering
    \includegraphics[width=\linewidth]{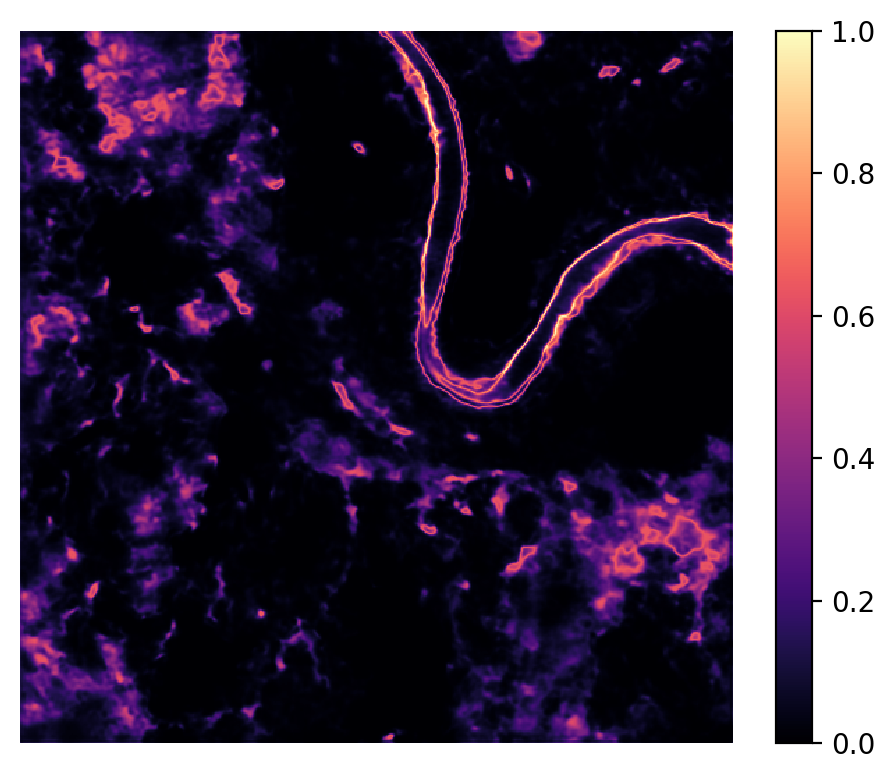}
    \caption{}
    \label{fig:scene1_unet_entropy}
\end{subfigure}
\hfill
\begin{subfigure}[t]{0.33\textwidth}
    \centering
    \includegraphics[width=\linewidth]{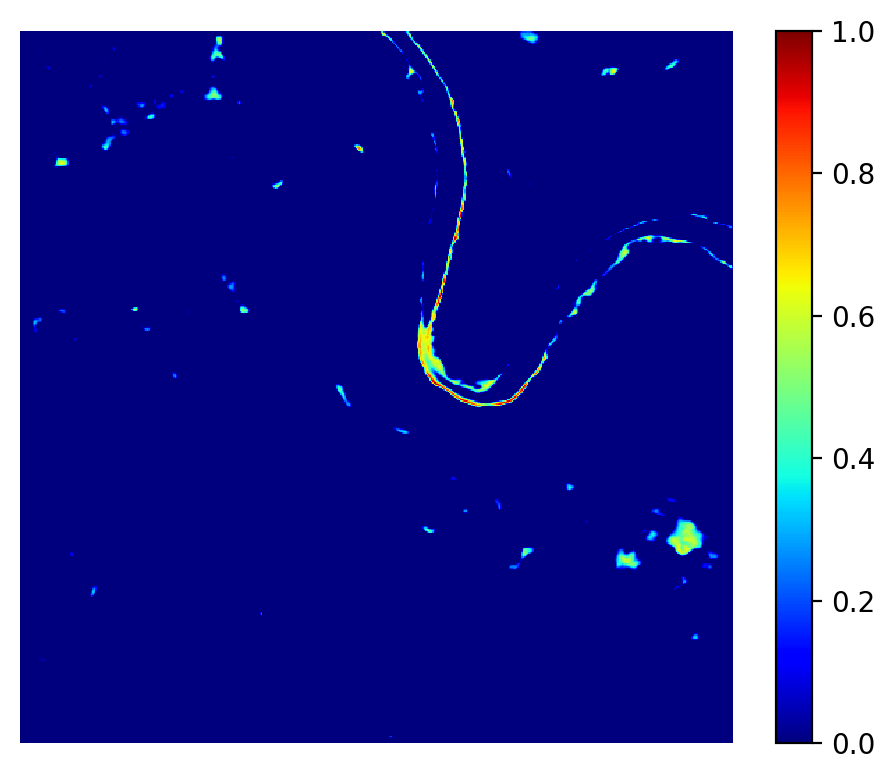}
    \caption{}
    \label{fig:scene1_unet_gradcam}
\end{subfigure}
\caption{Visualization of segmentation performance of the models on the Sen1Floods11 test data scene 1, (a) VV polarization SAR image, (b) flood ground truth, (c) permanent water ground truth, (d) flooded land prediction by SegFormer-b2, (e) entropy map of SegFormer-b2, (f) GRAD CAM activation map of SegFormer b2, (g) flooded land prediction by UNet, (h) entropy map of UNet, and (i) GRAD CAM activation map of UNet. The representative scene represents the severe monsoon flooding in Assam, India, on 12th August 2016.}
\label{fig:scene1}
\end{figure*}

\begin{figure*}[t]
\centering
\begin{subfigure}[t]{0.28\textwidth}
    \centering
    \includegraphics[width=\linewidth]{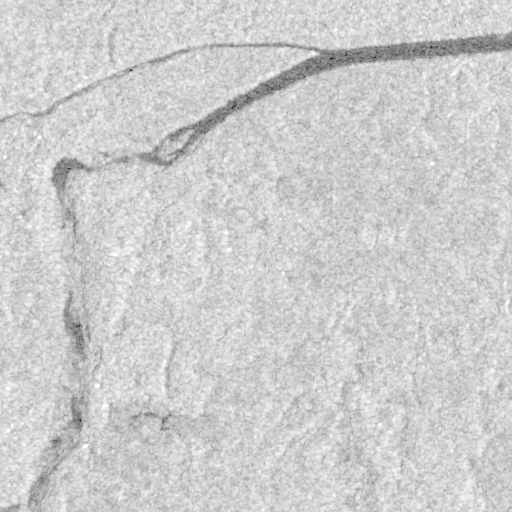}
    \caption{}
    \label{fig:scene2_vv}
\end{subfigure}
\hfill
\begin{subfigure}[t]{0.28\textwidth}
    \centering
    \hspace{-1.1cm}\includegraphics[width=\linewidth]{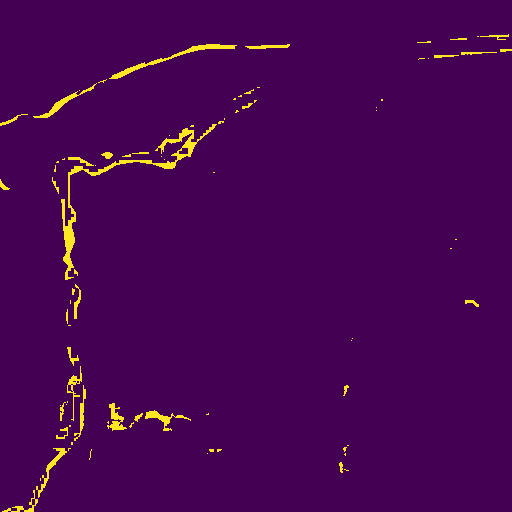}
    \caption{}
    \label{fig:scene2_flood_gt}
\end{subfigure}
\hfill
\begin{subfigure}[t]{0.28\textwidth}
    \centering
    \hspace{-1.1cm}\includegraphics[width=\linewidth]{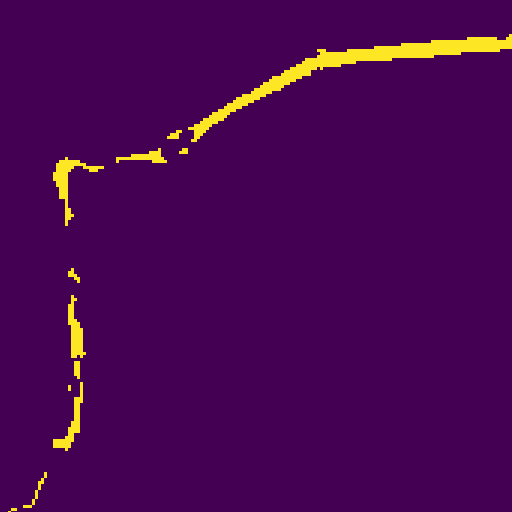}
    \caption{}
    \label{fig:scene2_water_gt}
\end{subfigure}
\vspace{0.3cm}
\begin{subfigure}[t]{0.28\textwidth}
    \centering
    \includegraphics[width=\linewidth]{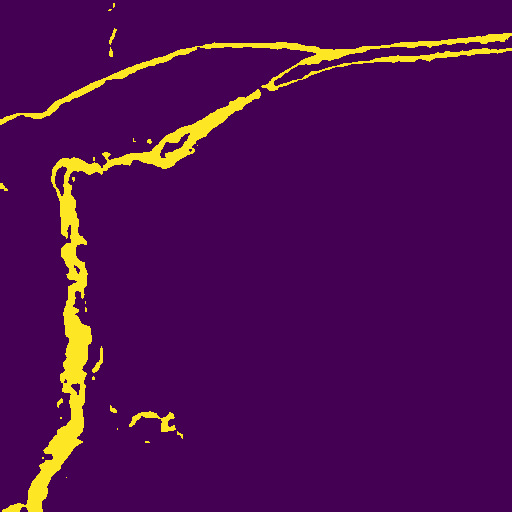}
    \caption{}
    \label{fig:scene2_segformer_prediction}
\end{subfigure}
\hfill
\begin{subfigure}[t]{0.33\textwidth}
    \centering
    \includegraphics[width=\linewidth]{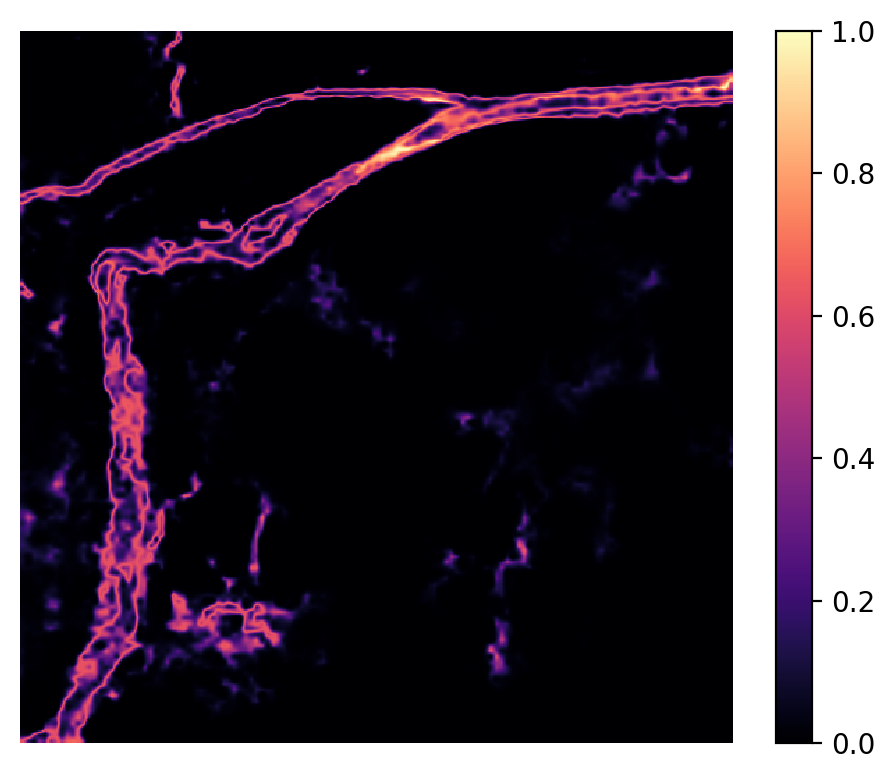}
    \caption{}
    \label{fig:scene2_segformer_entropy}
\end{subfigure}
\hfill
\begin{subfigure}[t]{0.33\textwidth}
    \centering
    \includegraphics[width=\linewidth]{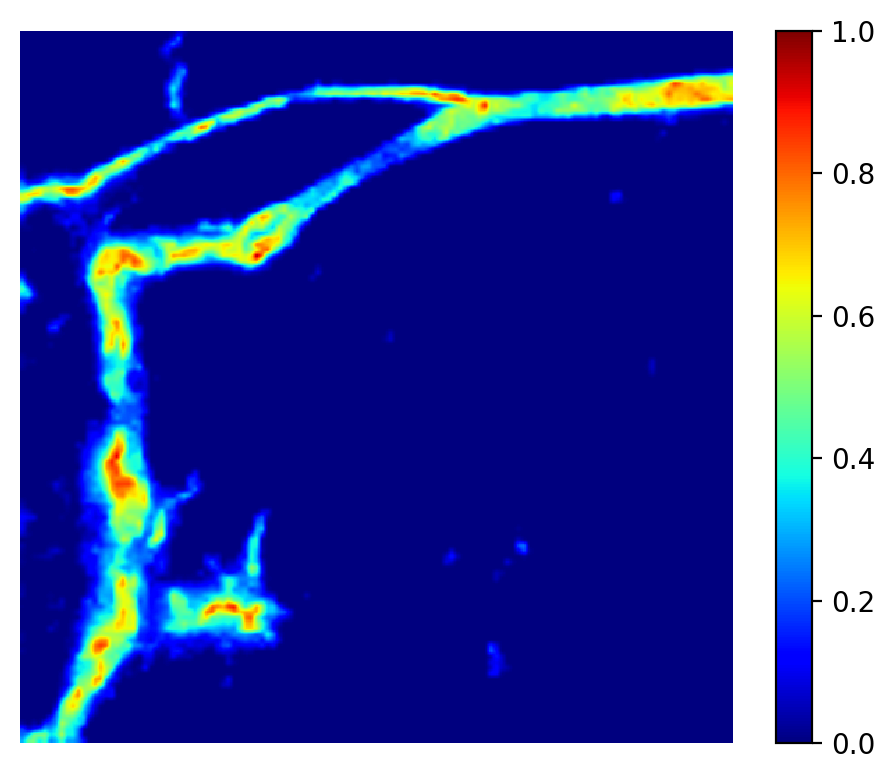}
    \caption{}
    \label{fig:scene2_segformer_gradcam}
\end{subfigure}

\vspace{0.3cm}

\begin{subfigure}[t]{0.28\textwidth}
    \centering
    \includegraphics[width=\linewidth]{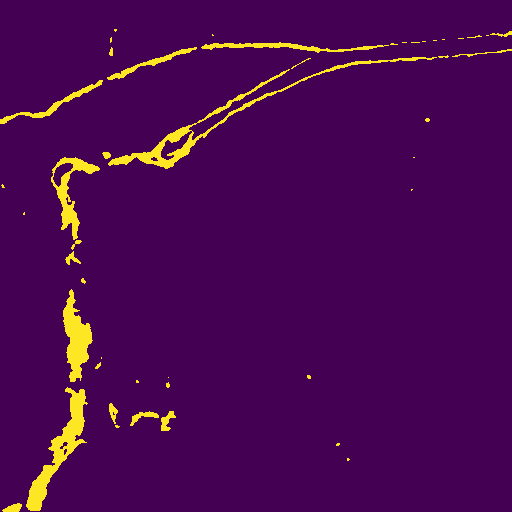}
    \caption{}
    \label{fig:scene2_unet_prediction}
\end{subfigure}
\hfill
\begin{subfigure}[t]{0.33\textwidth}
    \centering
    \includegraphics[width=\linewidth]{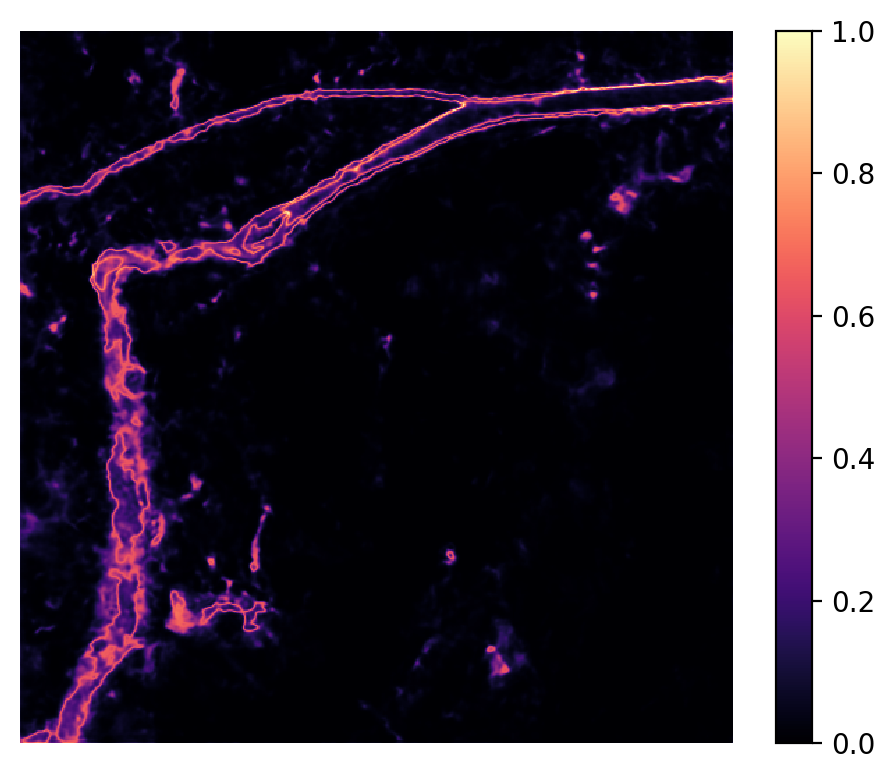}
    \caption{}
    \label{fig:scene2_unet_entropy}
\end{subfigure}
\hfill
\begin{subfigure}[t]{0.33\textwidth}
    \centering
    \includegraphics[width=\linewidth]{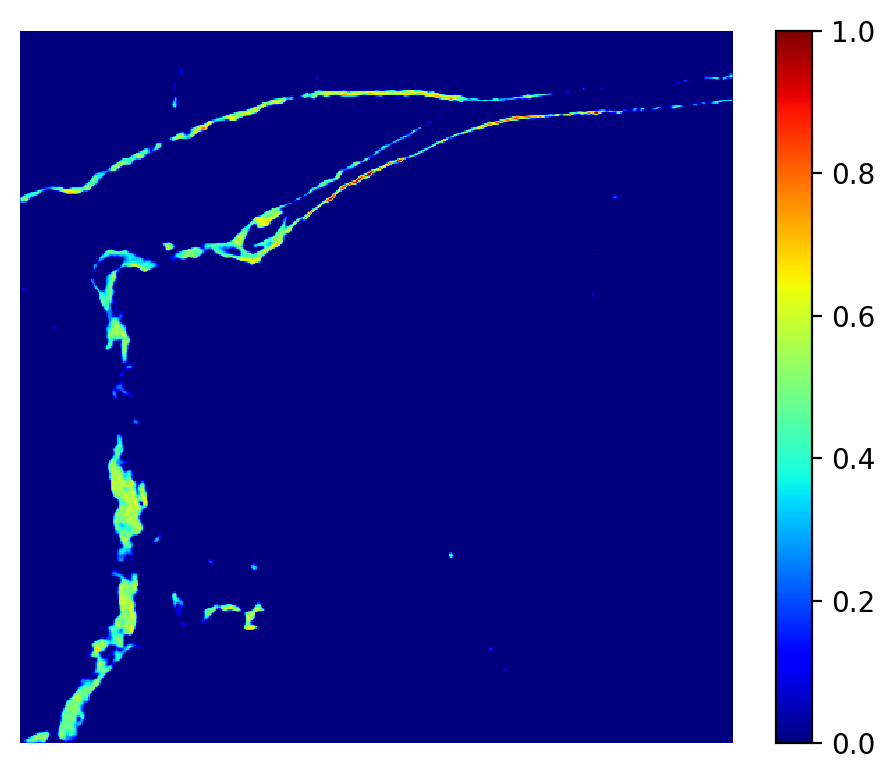}
    \caption{}
    \label{fig:scene2_unet_gradcam}
\end{subfigure}
\caption{Visualization of segmentation performance of the models on the Sen1Floods11 test data scene 2, (a) VV polarization SAR image, (b) flood ground truth, (c) permanent water ground truth, (d) flooded land prediction by SegFormer-b2, (e) entropy map of SegFormer-b2, (f) GRAD CAM activation map of SegFormer b2, (g) flooded land prediction by UNet, (h) entropy map of UNet, and (i) GRAD CAM activation map of UNet. The representative scene represents the severe monsoon flooding in Assam, India, on 12th August 2016.}
\label{fig:scene2}
\end{figure*}

The above results indicate that SegFormer-b2 is the most effective model for flood-class segmentation across both datasets. Therefore, the subsequent segmentation evaluation focuses on comparing the predictions of the best-performing vision model, segFormer-b2, with those of the standard UNet model. 

Figure 4 presents visual predictions from both SegFormer-b2 and UNet, along with Grad-CAM and entropy maps. In this figure, a region south of the Majuli River island in Assam, India, is shown as affected by monsoon flooding in 2016. Both models successfully segment the main flooded river course, and their predictions (Figures 4d and 4g) align closely with the flood ground truth in Figure 4(b). However, SegFormer-b2 predictions, as demonstrated in Figure 4(d), appear slightly more conservative along the channel margins and produce fewer scattered false positives in the surrounding region than UNet, which tends to overpredict isolated flood pixels, as seen in Figure 4(g). SegFormer-b2 successfully captures the river meander flooding, tracking the curving channels and adjacent inundated areas. It misses the large, flooded plain in the lower-left and many of the smaller scattered patches. However, it produces a coherent prediction that captures the main spatial structure of the flood event. In Figure 4(g), UNet fails almost entirely on this scene. Its prediction consists only of a few isolate, scattered regions in the image, completely missingr meander flooding. In the entropy maps shown in Figure 4(e) and Figure 4(h), SegFormer-b2 exhibits uncertainty along the water-land boundary, whereas UNet exhibits uncertainty across the wider scene, suggesting less confident discrimination between flooded and non-flooded surfaces. The Grad-CAM activation maps in Figures 4(f) and 4(i) further depict that SegFormer-b2's decision-making is strongly focused on the river itself, while UNet's activations are more fragmented. \par
The question worth addressing in future studies is whether substantial underprediction would be problematic in an operational flood monitoring context. The Grad-CAM maps reveal a striking contrast: in Figure 4(f), SegFormer-b2 generates strong high-intensity activations that closely track the river meanders, demonstrating that the transformer backbone clearly identifies flood-relevant features in this scene. On the other hand, in Figure 4(i), the UNet shows only very light, low-intensity activations, which explains its poor segmentation performance. 
\begin{table}[htbp]
\centering
\caption{Quantitative explainable Metrics for U-Net and SegFormer-b2 on the Sen1Floods11 Test Set}
\label{tab:xai_metrics}
\begin{tabular}{lccc}
\hline
\textbf{Metric} & \textbf{U-Net} & \textbf{SegFormer-b2} & \textbf{Better} \\
\hline
ECE & 0.0675 & 0.0694 & U-Net \\
Entropy--Error Corr.\ $r$ & 0.4408 & 0.4606 & SegFormer-b2 \\
AUDC & 0.0573 & 0.0477 & SegFormer-b2 \\
AUIC & 0.1494 & 0.1489 & U-Net \\
Pointing Game & 0.7407 & 0.6667 & U-Net \\
\hline
\end{tabular}
\end{table}

\begin{table}[htbp]
\centering
\caption{Spatial Entropy Distribution across semantic Zones on the Sen1Floods11 Test Set}
\label{tab:spatial_entropy}
\begin{tabular}{lcc}
\hline
\textbf{Spatial Zone} & \textbf{U-Net} & \textbf{SegFormer-b2} \\
\hline
Boundary & 0.2028 $\pm$ 0.0789 & 0.1846 $\pm$ 0.0848 \\
Interior Flood & 0.2535 $\pm$ 0.0897 & 0.2416 $\pm$ 0.1403 \\
Interior Non-Flood & 0.0535 $\pm$ 0.0611 & 0.0525 $\pm$ 0.0832 \\
\hline
\end{tabular}
\end{table}

\begin{figure*}[htbp]
\centering
\includegraphics[width=\textwidth]{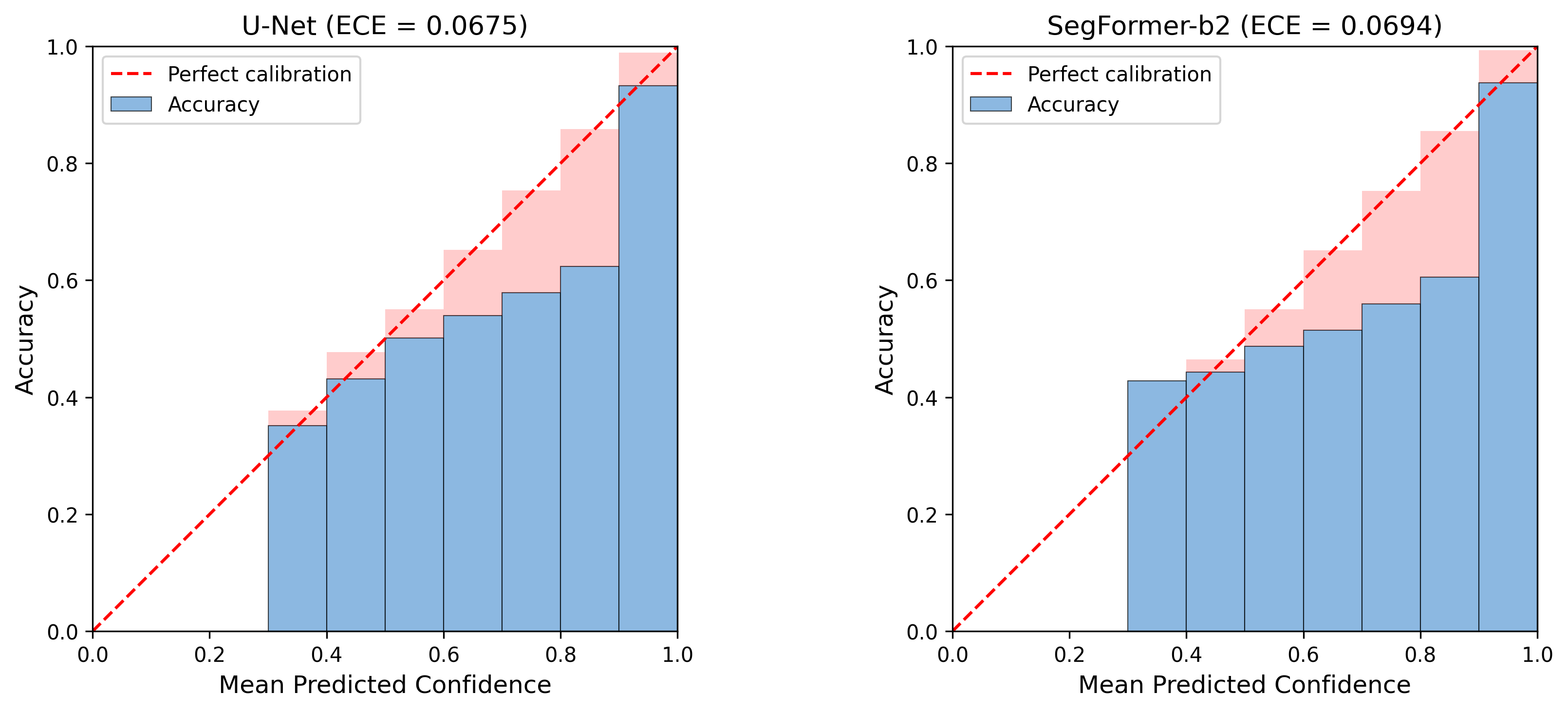}
\caption{Reliability diagrams for U-Net (left) and 
SegFormer-b2 (right) on the Sen1Floods11 test set. 
Both models are systematically overconfident, with accuracy bars 
falling below the perfect calibration diagonal. The red shaded 
regions indicate the calibration gap per bin.}
\label{fig:reliability}
\end{figure*}

\begin{figure*}[htbp]
\centering
\includegraphics[width=\textwidth]{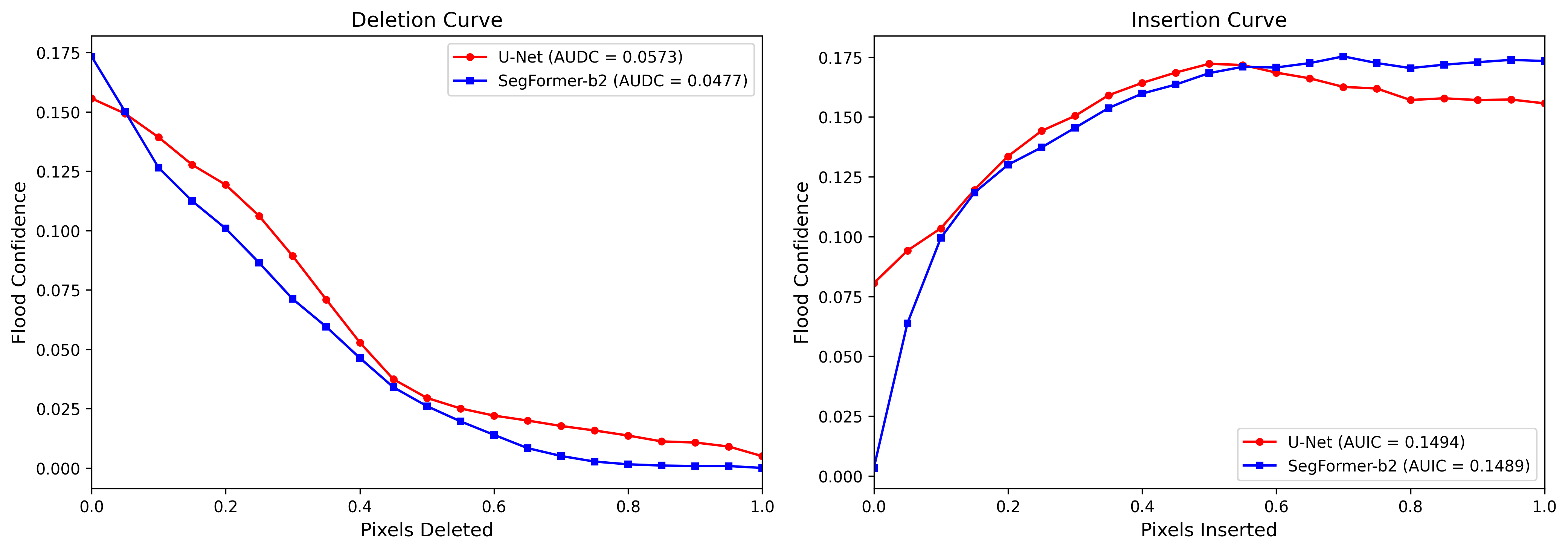}
\caption{Deletion and insertion faithfulness curves comparing U-Net 
and SegFormer-b2 Grad-CAM explanations on the Sen1Floods11 test set. 
SegFormer-b2 exhibits a steeper confidence decline as salient 
pixels are removed, indicating more faithful feature attribution. U-Net begins with a nontrivial baseline confidence, suggesting residual sensitivity to background texture, whereas SegFormer-b2 starts from near zero. Both models achieve comparable insertion performance.}
\label{fig:deletion_insertion}
\end{figure*}

Figure 5(a) shows a scene from Assam, India, dominated by a branching channel network with several finer tributaries. From Figures 5(d) to 5(g), it is understood that both models successfully detect the main horizontal channel along the top and the vertical channel on the right. However, the two predictions differ slightly in the characteristics detected. In Figure 5(d), SegFormer-b2 produces thicker, more continuous flood predictions and notably overpredicts at the channel junction in the lower portion of the scene, where it generates a large structure that is not present in the flood ground truth (Figure 5(b)). The UNet predictions shown in Figure 5(g) are thinner and more delicate, more closely resembling the flood ground truth. The entropy maps in Figure 5(e) and 5(h) indicate that both models are appropriately uncertain along the channel boundaries, with SegFormer-b2 producing slightly more spatially structured uncertainty. Grad-CAM maps shown in Figures 5(f) and 5(i) demonstrate the underlying cause of this difference. SegFormer-b2 produces high-intensity activations concentrated at the channel junction, which drive its over-prediction in that region, while UNet's weaker activations result in a more restrained but ultimately more accurate segmentation.
To complement the visual predictions, five quantitative XAI metrics are computed for U-Net and SegFormer-b2 on the Sen1Floods11 test set and are summarized in Tables V and VI. As seen in Table IV, U-Net achieves marginally better calibration as shown in the reliability diagrams in Figure 6, consistent with typical values reported for uncalibrated segmentation models \cite{breitling2025reliability, wang2025calibration}. 
The entropy–error correlation is also moderate for both architectures, indicating that predictive uncertainty serves as a meaningful proxy for segmentation errors, as reported in \cite{nasution2024interpretability, gonzalezjimenez2025wmh} for flood segmentation, respectively. 
Figure 7 presents a comparative evaluation of the two models in terms of deletion and insertion scores on the sen1floods11 test set. SegFormer-b2 achieves a lower deletion curve area, indicating that its Grad-CAM maps more accurately identify decision-relevant features, consistent with benchmarking studies showing that transformer attention maps capture more causally faithful attributions than CNN-based saliency methods \cite{petsiuk2018rise, abdelhalim2025xai}. The insertion scores are comparable across both models. The pointing game accuracy evaluated over the 27 flood-containing test chips of the Sen1Floods11 test set, is higher for U-Net (0.74 vs. 0.67). This indicates that its most salient Grad-CAM pixel falls within the flood ground truth more frequently. This aligns with the findings of \cite{azad2022contextual, gu2022convformer} on the trade-off between CNNs and transformer architectures. The transformer self-attention distributes feature importance globally, producing more faithful but spatially diffuse explanations, whereas CNN convolutions concentrate saliency locally, yielding more precise but less holistic attributions.
The spatial entropy analysis in Table VI demonstrates that flood interior regions exhibit higher uncertainty than boundary regions for both models, attributable to the inherent spectral ambiguity of shallow water and wet soil in SAR imagery \cite{zeineldin2025drowning}. SegFormer-b2 shows lower mean entropy across all zones but higher variance in flood interiors, suggesting scene-dependent overconfidence consistent with the qualitative observations. 
These complementary strengths suggest that, in operational flood monitoring, U-Net's explanations may be better suited to communicating localized flood boundaries, whereas SegFormer-b2's explanations better capture the overall relevance structure of flood-driving features.

\section{Conclusion}
This study investigated the performance of two image segmentation architectures, U-Net and SegFormer, specifically for detecting flooded land using Sentinel-1 SAR imagery. Six models were evaluated across two benchmark datasets, comprising two CNN-based architectures, three Vision Transformer variants (SegFormer-b0, SegFormer-b1, and SegFormer-b2), and additionally, DeepLab V3 within a multi-class segmentation framework that predicts 3 classes: flooded land, permanent water, and background/dry land. The experimental results demonstrated that SegFormer-b2 achieved the highest aggregate segmentation scores on both datasets; a paired per-scene analysis showed this advantage to be statistically significant on ETCI ($p = 0.016$, all seven test scenes), whereas after fine-tuning on Sen1Floods11 it fell within scene-to-scene variability, indicating that in-domain fine-tuning substantially narrows the gap between architectures. Flood detection proved substantially more challenging than permanent water segmentation in all models, with flood IoU scores roughly half those achieved for permanent water. This disparity reflects the inherent difficulty of identifying temporary flooding in SAR imagery, where flood boundaries are often ambiguous, and SAR backscatter intensities closely resemble those of permanent water bodies. All models exhibited a tendency toward higher recall than precision for the flood class, suggesting systematic overprediction of flood extent, a behavior that may be acceptable in operational early warning systems, where missed detections carry greater risk than false alarms. From the visual predictions, it was observed that while the U-Net captures the general flood extent, its predictions exhibit fragmented structures and irregular boundaries, particularly for elongated river patterns. This behavior indicates dependence on local features. In contrast, the SegFormer-b2 model produces spatially coherent predictions and preserves the continuity of thin flood structures, benefiting from its ability to model long-range dependencies. This behavior can be attributed to the self-attention mechanism in the transformer encoder, which assimilates information across the entire spatial extent of the feature map. In contrast, U-Net's reliance on local convolutional kernels limits its effective receptive field, resulting in fragmented predictions for spatially extended features but finer sensitivity along class boundaries where local context is sufficient.
The entropy and Grad-CAM activation map visualization revealed important differences in how the architectures process flood-relevant characteristics. SegFormer-b2 produced more spatially focused and discriminative Grad-CAM activations compared to UNet, confirming that the transformer backbone learns stronger feature representations for flood detection. However, the entropy analysis revealed a notable trade-off: SegFormer-b2 exhibited overconfidence in certain scenes, producing low entropy values even in incorrectly predicted regions, while UNet generated more informative uncertainty estimates along boundary regions. 
The quantitative XAI evaluation reinforced these qualitative observations: SegFormer-b2 achieved more faithful explanations and a slightly stronger entropy–error correlation, while U-Net demonstrated more spatially precise explanations, a higher pointing game score, and marginally better calibration. The spatial entropy analysis further revealed that flood interior regions exhibit higher uncertainty than boundary regions for both models, reflecting the inherent spectral ambiguity of shallow water and wet soil in SAR imagery. SegFormer-b2 also showed greater variance in flood-interior confidence, demonstrating scene-dependent overconfidence. These complementary explainability characteristics have direct practical implications: SegFormer-b2's more faithful explanations are better suited to automated systems that rely on saliency maps for feature prioritization, whereas U-Net's spatially precise and better-calibrated uncertainty estimates are more appropriate for communicating flood-risk boundaries to emergency responders. Several directions for future work emerge from this study. First, the overconfidence observed in SegFormer-b2 motivates exploring calibration techniques such as temperature scaling, Monte Carlo dropout, or deep ensemble methods to produce more reliable uncertainty estimates for transformer-based architectures. Second, incorporating multi-temporal SAR data, in which pre-flood and post-flood images are jointly analyzed, could help models better distinguish temporary flooding from permanent water bodies. Third, integrating additional data sources, such as digital elevation models or optical imagery, could provide complementary information to improve segmentation accuracy in challenging scenes. Finally, evaluating these architectures on higher-resolution SAR data or in near-real-time operational pipelines would help assess their practical readiness for deployment in disaster response scenarios.



\bibliographystyle{IEEEtran}
\bibliography{references}
\end{document}